\documentclass[conference]{IEEEtran}
\IEEEoverridecommandlockouts
\usepackage{cite}
\usepackage{amsmath,amssymb,amsfonts}
\usepackage{algorithmic}
\usepackage{graphicx}
\usepackage{textcomp}
\usepackage{xcolor}
\usepackage{subcaption}
\usepackage[most]{tcolorbox}
\usepackage{amsmath}
\usepackage{lmodern}
\usepackage{geometry}
\usepackage{booktabs}
\usepackage{multirow}
\usepackage{longtable}
\usepackage{lscape}
\usepackage{soul}
\usepackage{array}
\usepackage{booktabs} %
\usepackage{makecell}

\def\BibTeX{{\rm B\kern-.05em{\sc i\kern-.025em b}\kern-.08em
    T\kern-.1667em\lower.7ex\hbox{E}\kern-.125emX}}
\begin{document}

\title{On-Device Language Models for Privacy-Preserving Stress Prediction: A Multimodal Evaluation on Mobile Health}

\author{
    \IEEEauthorblockN{
        Ibukunoluwa Soyebo\textsuperscript{1}, Alyssa Donawa\textsuperscript{1}, Rodrigo Aguilar Barrios\textsuperscript{1}, Brice Patchou\textsuperscript{2}, Corey E. Baker\textsuperscript{1,2}
    }
    \IEEEauthorblockA{
        \textit{
        \textsuperscript{1}Ming Hsieh Department of Electrical \& Computer Engineering,
        \textsuperscript{2}Thomas Lord Department of Computer Science
        } \\
        University of Southern California \\
        \{isoyebo, donawa, aguilarb, patchou, c.baker\}@usc.edu
    }
}

\maketitle

\begin{abstract}
Stress is a pervasive determinant of mental health and a key target for mobile health interventions.
On-device language models (ODLMs) offer privacy-preserving inference without cloud dependency, yet their feasibility for health prediction under mobile resource constraints remains underexplored.
We evaluate ODLMs for multi-modal stress prediction using zero-shot prompting, measuring predictive accuracy alongside latency and throughput.
Our results show that objective sensor features marginally outperform subjective self-reports on average, and that lightweight sub-2B models achieve low latency with predictable resource usage.
Our findings highlight both the promise and the practical constraints of ODLMs for mobile mental health.

\end{abstract}

\section{Introduction}
Stress is a routinely occurring condition in everyday life that negatively influences cognitive performance, emotional regulation, and long-term physical and mental health~\cite{dongre2024physiology, liu2023chatcounselor, mcewen_revisiting_2020}.
Prolonged and chronic stress is associated with adverse health and mental health outcomes, motivating the development of mobile health (mHealth) systems capable of detecting and predicting stress in real-world settings~\cite{ma2024understanding, mcewen_revisiting_2020}.
Advances in wearable sensing and smartphone-based data collection have enabled continuous capture of multimodal physiological and behavioral signals, including: heart rate variability, activity levels, sleep patterns, and self-reported affect\cite{thambawita2020pmdata, xu2022globem}.
These sensing capabilities have supported machine learning (ML) approaches for stress prediction from time-series data in ubiquitous computing environments\cite{wang2024efficient}.

Most traditional stress detection systems rely on task-specific supervised models trained directly on structured numerical features \cite{hua2025charting}.
While effective, these models are typically limited to predefined prediction objectives and require retraining for new tasks or data configurations.
In parallel, large language models (LLMs) have demonstrated flexible reasoning abilities across a range of structured and semi-structured inputs\cite{thapa2025stressllm, haque2024state}.
Recent work suggests that language models can interpret serialized time-series data using zero-shot prompting, raising the possibility of more generalizable inference pipelines that reduce task-specific engineering\cite{kim2024healthllm}.

However, deploying LLM-based inference in
mHealth contexts presents significant challenges.
Cloud-based models introduce privacy risks when handling sensitive physiological and behavioral data, and network latency can hinder real-time feedback.
Recent advances in scaling down LLMs through quantization\cite{bondarenko2021wquantization}, pruning\cite{li2020pruning}, and knowledge distillation \cite{ji2021distribution} have enabled small language models to operate directly on edge devices.
On-device language models (ODLMs) provide stronger privacy guarantees and reduced latency, but their capacity for reasoning over multimodal health time-series data under resource constraints remains insufficiently understood.

In this work, we investigate the feasibility of using ODLMs for stress prediction from combined objective (sensor-derived) and subjective (self-reported) time-series data.
Rather than relying on fine-tuning, we evaluate structured zero-shot prompting approaches that vary health data modality and temporal context serialization.
We systematically measure predictive accuracy alongside device-level metrics, including inference latency and throughput, to characterize the trade-offs between model capability and edge-device constraints.

\noindent\textbf{Our contributions are:}
(1) An empirical evaluation of ODLMs for multimodal stress event prediction using structured zero-shot prompting and
(2) a systematic analysis of modality composition and temporal serialization strategies for time-series reasoning under edge constraints.
By examining predictive capability, resource efficiency, and deployment feasibility, this work advances understanding of how ODLMs can support real-time, privacy-sensitive stress monitoring in ubiquitous computing systems.

\section{Related Works}
Recent work has explored the use of LLMs for structured health event prediction tasks.
Kim et al. \cite{kim2024healthllm} evaluated the ability of popular LLMs to infer subjective health outcomes from objective and self-reported data across multiple datasets.
Their study compared prompting strategies including zero-shot, few-shot, chain-of-thought, self-consistency, and fine-tuning, demonstrating that explicitly defining task context substantially improves predictive accuracy.
Wang et al. \cite{wang2024efficient} extended this line of inquiry to ODLMs, evaluating zero-shot prompting within the HealthLLM framework using the PMData dataset \cite{thambawita2020pmdata}.
They reported that smaller models such as TinyLlama-1.1B offered comparable inference accuracy to larger cloud-based models while offering advantages in latency and privacy.

Parallel efforts have explored LLMs for mental health assessment and counseling applications. Wei et al. \cite{wei2025mophes} introduced MoPHES, a mobile framework that fine-tunes compact LLMs for multi-turn mental health support, demonstrating that domain adaptation can enable competitive on-device performance.
Liu et al. \cite{liu2023chatcounselor} proposed ChatCounselor, trained on professionally curated counseling dialogues, showing that high-quality domain data substantially improves response quality.
Zheng et al. \cite{zheng2023building} addressed data scalability by synthesizing large-scale emotional support dialogues to improve empathy and contextual grounding.
While these systems demonstrate the promise of LLM-based mental health tools, they largely rely on fine-tuning and curated conversational corpora rather than structured multimodal physiological inputs.
Furthermore, many approaches prioritize dialogue generation quality over deployment trade-offs such as latency and memory footprint.

\section{Methodology}
\begin{table*}[t]
\scriptsize
\centering
\caption{An example of constructed zero-shot prompts}
\label{tab:zero_shot_prompt}
\renewcommand{\arraystretch}{1.4}

\begin{tabular}{p{0.22\textwidth} p{0.74\textwidth}}
\hline
\textbf{Context} & \textbf{Prompt} \\
\hline

\textbf{System Prompt} &
You are a personalized healthcare agent trained to predict a stress score, which ranges from 1 to 5 based on physiological data and user information. \\
\hline

\textbf{User Context} &
Given the user’s profile as age: \{48\}-year-old, sex: \{male\} and height: \{195\} cm \\
\hline

\textbf{Temporal Context} &
The recent data from the past 2-weeks \{14 days\} show: \\
\hline

\textbf{Health Data (NLS)} &
\begin{tabular}[t]{@{}p{0.36\textwidth} p{0.36\textwidth}@{}}
\textbf{Objective Data} &
\textbf{Subjective Data} \\
\hline
\raggedright
\textbf{Steps}: \{1476.0, 4809.0, \ldots, NaN\} \\
\textbf{Burned calories}: \{169.0, 419.0, \ldots, NaN\} \\
\textbf{Resting heart rate}: \{53.24, \ldots, 51.40\} \\
\textbf{Sleep minutes}: \{110.0, 524.0, \ldots, 481.0\} &
\raggedright
\textbf{Mood} (Out of 5): \{3.0, 4.0, \ldots, NaN\} \\
\textbf{Sleep Quality} (Out of 5): \{1.0, \ldots, NaN\} \\
\textbf{Fatigue} (Out of 5): \{1.0, 5.0, \ldots, NaN\} \\
\end{tabular}
\\
\hline

\textbf{Label Conditioning} &
Score 3 is normal; 1–2 are below normal; 4–5 are above normal. \\
\hline

\textbf{Answer prompt} &
The response should be a single discrete integer value. Use the following format to produce output:
“The predicted stress level is [Your response here]” \\
\hline
\end{tabular}
\end{table*}

To evaluate selected state-of-the-art ODLMs, we integrated LLMs into a research-based platform, ``Assuage'' with support for offline storage and data synchronization~\cite{donawa2024designing}.
Experiments were conducted on both iOS and iPadOS devices, all running version 18.5 of their respective operating systems.
\subsection{Model Selection}
Our experimental design combines methods from prior work.
Like \cite{nissen2025medicine}, we make use of MLXExamples to facilitate deployment, download, and interfacing with ODLMs available on HuggingFace.
Model sizes are classified as small ([1, 3) billion), medium ([3, 7) billion), and large ([7, $\infty$) billion)~\cite{nissen2025medicine}.
Our selected
ODLMs are as follows. \textbf{Qwen3-0.6B} and \textbf{Qwen3-1.7B} are quantized variants of the Qwen3 family from Alibaba, with approximately 0.6B and 1.7B parameters and on-device storage footprints of \emph{0.33}GB and \emph{0.94}GB, respectively.
\textbf{Granite-3.3-2b} is a 2-billion parameter instruction-tuned model from IBM, occupying \emph{1.33}GB on device.

\subsection{Prompting Strategies}
Our evaluation follows established prompt engineering practices, including zero-shot prompting, temporal context prompting, user context, label conditioning, and structured answer formatting~\cite{gruver2023large, jin2023time, belyaeva2023multimodal, kim2024healthllm, thapa2025stressllm}.
\textbf{Temporal Context Prompting}: To present time-series data to the language model, we adopt temporal context prompting. Prior work categorizes temporal context representations into three classes: Natural Language String (NLS)~\cite{gruver2023large}, Statistical Summary (SS)~\cite{jin2023time}, and Modality-specific Encoding~\cite{belyaeva2023multimodal}. In this study, we employ the first two representations.
NLS representation presents time-series data as ordered string sequences, using NaN for missing entries.
SS representation encodes time-series data using aggregate statistics (mean, standard deviation, minimum, and maximum)\footnote{\scriptsize Sleep in Hours: avg = 6.03, std = 2.53, min = 0.97, max = 9.83.}.
To ensure fair comparison across models, prompt templates were standardized and not optimized per model. All ODLMs received identical system instructions, temporal context formatting, and answer constraints.
We intentionally avoided model-specific prompt tuning to isolate the effects of model architecture, scale, and quantization on stress prediction performance.
While prompt optimization may improve individual model accuracy, our goal was to evaluate robustness under a consistent prompting framework.
\subsection{Dataset}
PMData is a five-month life-logging dataset comprising subjective and objective time-series health data from 16 participants~\cite{thambawita2020pmdata}. Objective data were collected via a Fitbit Versa 2 and include multi-rate measurements such as steps, calories, heart rate, and sleep. Subjective data were self-reported through the PMSys mobile app and Google Forms, capturing metrics including sleep quality, fatigue, mood, and readiness.

\section{Experiment}

\begin{figure}
    \centering
    \includegraphics[width=\linewidth, height=3.4in]{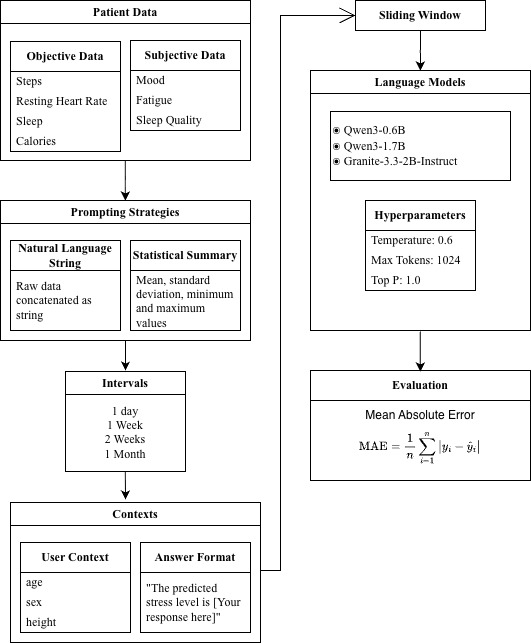}
    \caption{Experiment workflow}
    \label{fig:experiment}
\end{figure}

Figure~\ref{fig:experiment} illustrates the high-level workflow of our experimental pipeline.
Prior work evaluating language models for stress prediction on the PMData dataset commonly adopts the original prompt proposed by \cite{kim2024healthllm}, which combines objective measures
with the subjective Mood score.
In contrast, we decompose this prompt and extend it by incorporating additional metrics available in the dataset, allowing us to isolate the contribution of objective and subjective signals to stress prediction.

Table~\ref{tab:zero_shot_prompt} presents a sample of our prompt design, illustrating the three prompt variants developed in this study.
The first consists of objective features only and mirrors the default prompt used in prior work, excluding the subjective Mood score.
The second consists exclusively of subjective features.
Each subjective metric is contextualized relative to its upper bound, and the semantic meaning of the scale is explicitly defined by specifying which values correspond to low, normal, and high levels.
The third prompt combines both objective and subjective features.

We evaluate these models with prompts that capture four time intervals: 1 day (1D), 1 week (1W), 2 weeks (2W) and 1 month (1M), using both NLS and SS prompting strategies.
Each experimental instance is defined by a reference date and its associated retrospective temporal window. The prompt conditions the language model on data collected during this window, including subjective and objective signals recorded on the reference date, and tasks the model with nowcasting the patient’s stress level for that date.
For our baseline comparison, we evaluated several traditional ML models, including Decision Trees, Random Forests (RF), Support Vector Machines (SVM), Gradient Boosting, k-Nearest Neighbors (kNN), Naive Bayes, and Logistic Regression, using a 70:30 train-test split.

\begin{table*}[t]
\centering
\caption{MAE comparison using Traditional ML Baselines}
\label{tab:baseline}
\resizebox{\textwidth}{!}{%
\begin{tabular}{lcccc|cccc|cccc}
\toprule
Model &
1D & 1W & 2W & 1M &
1D & 1W & 2W & 1M &
1D & 1W & 2W & 1M \\
\cmidrule(lr){2-5}
\cmidrule(lr){6-9}
\cmidrule(lr){10-13}
& \multicolumn{4}{c}{Objective $\downarrow$}
& \multicolumn{4}{c}{Subjective $\downarrow$}
& \multicolumn{4}{c}{Objective + Subjective $\downarrow$}\\
\midrule

Decision Trees &
0.96$\pm$0.17 & 0.96$\pm$0.20 & 0.82$\pm$0.27 & 0.91$\pm$0.18 &
0.69$\pm$0.15 & 0.65$\pm$0.20 & 0.71$\pm$0.23 & 0.77$\pm$0.19 &
0.68$\pm$0.13 & 0.69$\pm$0.25 & 0.72$\pm$0.19 & 0.70$\pm$0.10 \\

RF &
0.69$\pm$0.17 & \textbf{0.58$\pm$0.24} & \textbf{0.55$\pm$0.22} & \textbf{0.52$\pm$0.20} &
0.58$\pm$0.17 & 0.53$\pm$0.20 & \textbf{0.52$\pm$0.19} & \textbf{0.52$\pm$0.19} &
0.52$\pm$0.21 & 0.53$\pm$0.19 & \textbf{0.52$\pm$0.21} & \textbf{0.50$\pm$0.20} \\

SVM &
\textbf{0.56$\pm$0.13} & 0.61$\pm$0.12 & 0.59$\pm$0.11 & 0.54$\pm$0.16 &
\textbf{0.52$\pm$0.18} & \textbf{0.52$\pm$0.20} & 0.57$\pm$0.08 & 0.54$\pm$0.12 &
\textbf{0.52$\pm$0.19} & \textbf{0.50$\pm$0.19} & 0.53$\pm$0.15 & 0.53$\pm$0.15 \\

Gradient Boosting &
0.91$\pm$0.20 & 0.77$\pm$0.20 & 0.70$\pm$0.19 & 0.71$\pm$0.21 &
0.63$\pm$0.15 & 0.62$\pm$0.23 & 0.59$\pm$0.22 & 0.57$\pm$0.21 &
0.55$\pm$0.18 & 0.56$\pm$0.23 & 0.53$\pm$0.24 & 0.54$\pm$0.22 \\

kNN &
0.90$\pm$0.26 & 0.82$\pm$0.22 & 0.83$\pm$0.18 & 0.92$\pm$0.23 &
0.61$\pm$0.18 & 0.69$\pm$0.07 & 0.73$\pm$0.09 & 0.74$\pm$0.11 &
0.66$\pm$0.24 & 0.73$\pm$0.09 & 0.72$\pm$0.12 & 0.77$\pm$0.11 \\

Na\"ive Bayes &
0.71$\pm$0.10 & 0.95$\pm$0.14 & 0.95$\pm$0.13 & 1.04$\pm$0.11 &
0.83$\pm$0.17 & 0.72$\pm$0.26 & 0.72$\pm$0.31 & 0.70$\pm$0.34 &
0.73$\pm$0.10 & 0.73$\pm$0.20 & 0.78$\pm$0.17 & 0.93$\pm$0.11 \\

Logistic Regression &
1.35$\pm$0.62 & 1.34$\pm$0.63 & 1.42$\pm$0.56 & 1.61$\pm$0.46 &
1.23$\pm$0.59 & 1.02$\pm$0.33 & 0.91$\pm$0.23 & 0.91$\pm$0.14 &
1.19$\pm$0.55 & 0.89$\pm$0.25 & 0.97$\pm$0.26 & 1.04$\pm$0.16 \\

\bottomrule
\end{tabular}
}
\end{table*}

\begin{table*}[t]
\centering
\caption{Model Performance - Natural Language String}
\label{tab:nlsMae}
\resizebox{\textwidth}{!}{%
\begin{tabular}{lcccc|cccc|cccc}
\toprule
\textbf{Model} &
\textbf{1D} & \textbf{1W} & \textbf{2W} & \textbf{1M} &
\textbf{1D} & \textbf{1W} & \textbf{2W} & \textbf{1M} &
\textbf{1D} & \textbf{1W} & \textbf{2W} & \textbf{1M} \\

\cmidrule(lr){2-5}
\cmidrule(lr){6-9}
\cmidrule(lr){10-13} &
\multicolumn{4}{c}{\textbf{Objective \boldmath$\downarrow$}} &
\multicolumn{4}{c}{\textbf{Subjective \boldmath$\downarrow$}} &
\multicolumn{4}{c}{\textbf{Objective + Subjective \boldmath$\downarrow$}} \\
\midrule

Qwen3-0.6B &
\textbf{0.57 ± 0.36} & \textbf{0.56 ± 0.31} & \textbf{0.60 ± 0.31} & \textbf{0.57 ± 0.32} &
\textbf{0.53 ± 0.34} & \textbf{0.57 ± 0.35} & \textbf{0.54 ± 0.33} & \textbf{0.51 ± 0.35} &
\textbf{0.55 ± 0.35} & \textbf{0.55 ± 0.32} & \textbf{0.56 ± 0.34} & \textbf{0.55 ± 0.35} \\

Qwen3-1.7B &
0.58 ± 0.38 & 0.58 ± 0.29 & 0.63 ± 0.37 & 0.61 ± 0.40 &
0.59 ± 0.37 & 0.59 ± 0.33 & 0.61 ± 0.37 & 0.59 ± 0.40 &
0.59 ± 0.41 & 0.59 ± 0.38 & 0.58 ± 0.38 & 0.58 ± 0.43 \\

Granite-3.3-2b  &
0.62 ± 0.30 & 0.73 ± 0.21 & 0.70 ± 0.23 & 0.70 ± 0.22 &
0.85 ± 0.20 & 0.80 ± 0.15 & 0.86 ± 0.19 & 0.73 ± 0.22 &
0.84 ± 0.11 & 0.72 ± 0.16 & 0.74 ± 0.17 & 0.64 ± 0.24 \\

\bottomrule
\end{tabular}
}
\end{table*}

\begin{table*}[t]
\centering
\caption{Model Performance - Statistical Summary}
\label{tab:ssMAE}
\resizebox{\textwidth}{!}{%
\begin{tabular}{lcccc|cccc|cccc}
\toprule
\textbf{Model} &
\textbf{1D} & \textbf{1W} & \textbf{2W} & \textbf{1M} &
\textbf{1D} & \textbf{1W} & \textbf{2W} & \textbf{1M} &
\textbf{1D} & \textbf{1W} & \textbf{2W} & \textbf{1M} \\
\cmidrule(lr){2-5}
\cmidrule(lr){6-9}
\cmidrule(lr){10-13} &
\multicolumn{4}{c}{\textbf{Objective \boldmath$\downarrow$}} &
\multicolumn{4}{c}{\textbf{Subjective \boldmath$\downarrow$}} &
\multicolumn{4}{c}{\textbf{Objective + Subjective \boldmath$\downarrow$}} \\
\midrule
Qwen3-0.6B &
0.55 ± 0.34 & 0.56 ± 0.27 & 0.58 ± 0.28 & 0.51 ± 0.28 &
\textbf{0.52 ± 0.34} & \textbf{0.52 ± 0.29} & \textbf{0.51 ± 0.29} & \textbf{0.48 ± 0.28} &
0.54 ± 0.34 & 0.53 ± 0.30 & 0.53 ± 0.29 & 0.49 ± 0.29 \\

Qwen3-1.7B &
\textbf{0.43 ± 0.24} & \textbf{0.52 ± 0.30} & \textbf{0.55 ± 0.28} & \textbf{0.51 ± 0.27} &
\textbf{0.52 ± 0.34} & 0.54 ± 0.30 & 0.53 ± 0.30 & 0.49 ± 0.29 &
\textbf{0.54 ± 0.32} & \textbf{0.52 ± 0.29} & \textbf{0.51 ± 0.29} & \textbf{0.48 ± 0.29} \\

Granite-3.3-2b &
0.54 ± 0.32 & 0.59 ± 0.27 & 0.57 ± 0.25 & 0.52 ± 0.28 &
0.72 ± 0.31 & 0.66 ± 0.26 & 0.67 ± 0.22 & 0.58 ± 0.22 &
0.63 ± 0.28 & 0.62 ± 0.25 & 0.59 ± 0.24 & 0.57 ± 0.26  \\

\bottomrule
\end{tabular}
}
\end{table*}

\begin{table*}[t]
\centering
\caption{Average Prompt Token Count}
\label{tab:prompt-token-count-comparison}
\resizebox{\textwidth}{!}{%
\begin{tabular}{lcccc|cccc|cccc}
\toprule
Prompting Strategy &
1D & 1W & 2W & 1M &
1D & 1W & 2W & 1M  &
1D & 1W & 2W & 1M \\
\cmidrule(lr){2-5}
\cmidrule(lr){6-9}
\cmidrule(lr){10-13}
& \multicolumn{4}{c}{Objective}
& \multicolumn{4}{c}{Subjective}
& \multicolumn{4}{c}{Objective + Subjective}\\
\midrule
Natural Language String
& 209.08 & 351.75 & 525.64 & 908.56
& 192.70 & 241.51 & 301.26 & 416.03
& 247.30 & 429.80 & 655.98 & 1161.47\\

Statistical Summary
& 343.09 & 349.43 & 356.97 & 349.62
& 271.36 & 280.70 & 288.66 & 280.87
& 447.93 & 460.93 & 468.72 & 461.37 \\
\bottomrule
\end{tabular}
}
\end{table*}

\begin{figure*}[t]
\centering
\begin{subfigure}[b]{0.48\textwidth}
\centering
\includegraphics[width=\linewidth]{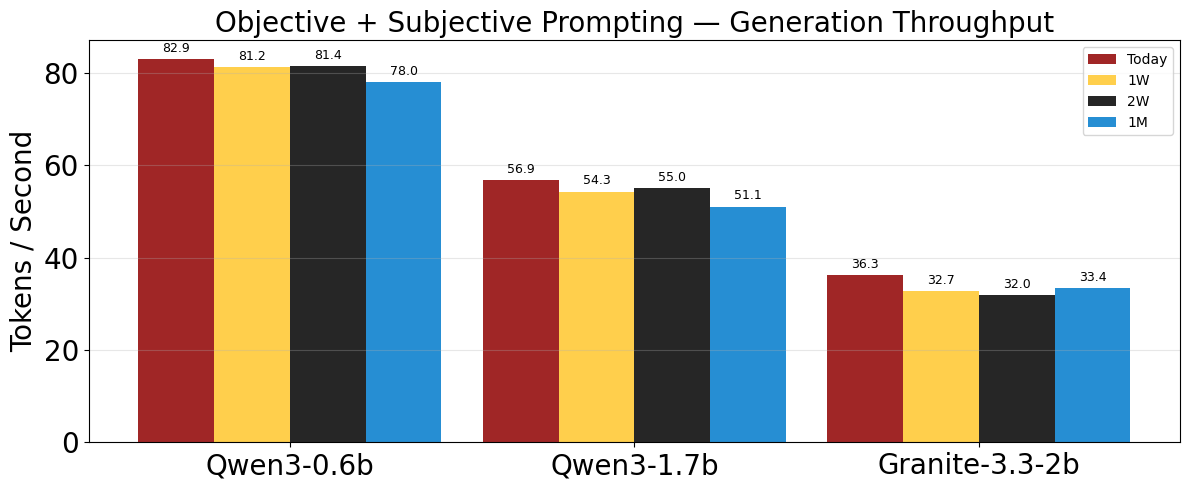}
\caption{Throughput}
\label{fig:on-device-eval-a}
\end{subfigure}
\hfill
\begin{subfigure}[b]{0.48\textwidth}
\centering
\includegraphics[width=\linewidth]{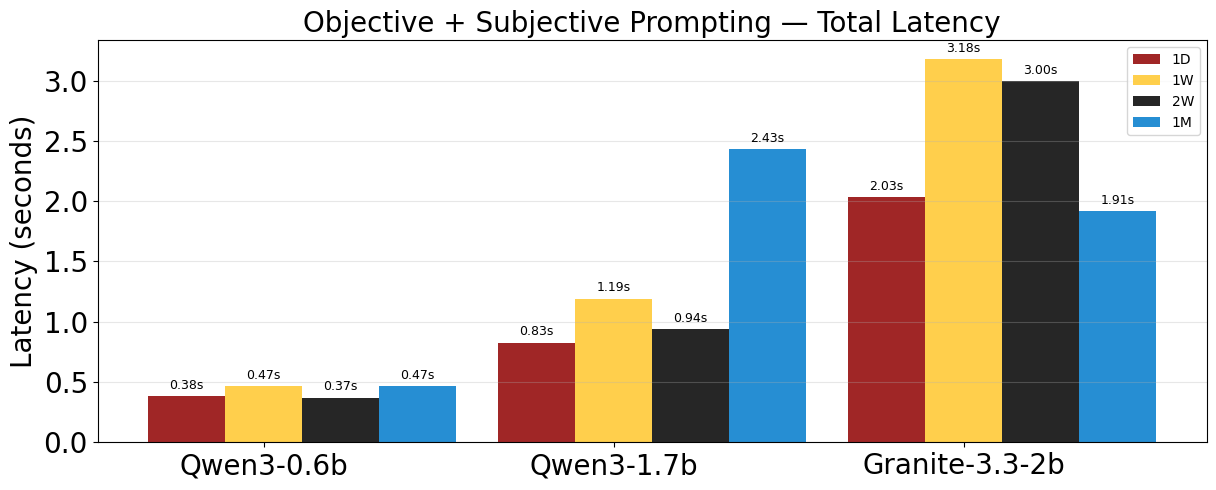}
\caption{Latency}
\label{fig:on-device-eval-b}
\end{subfigure}

\caption{Device Performance Metrics}
\label{fig:on-device-eval}
\end{figure*}

\section{Results}
Model performance was evaluated using Mean Absolute Error (MAE) between participants' reported stress values and the language models' predictions, where lower MAE values indicate better agreement with the ground truth.

\subsection{Traditional Machine Learning Baselines}
Table~\ref{tab:baseline} reports the MAE of conventional supervised models across objective-only, subjective-only, and combined feature configurations for the four time intervals.
Overall, RF and SVM consistently achieve the strongest performance across the evaluated settings.
For objective features, SVM attains the lowest MAE at the interval 1D  (0.56), while RF
performs best at 1W (0.58), 2W (0.55), and 1M (0.52).
For subjective features, SVM achieves the lowest MAE at 1D (0.52) and 1W (0.52), whereas RF performs best at 2W (0.52) and 1M (0.52).
For the combined objective and subjective feature set, SVM achieves the lowest MAE at 1D (0.52) and 1W (0.50), while RF obtains the best results at 2W (0.52) and 1M (0.50).
Across all time intervals, the combined feature configuration generally yields the lowest errors, indicating that integrating objective and subjective signals provides complementary information that improves predictive performance.

\subsection{Model Accuracy: Natural Language String}
Table~\ref{tab:nlsMae} summarizes model performance under the NLS prompting strategy.
Although longer temporal windows produce substantially longer prompts, particularly for the 1M interval (Table~\ref{tab:prompt-token-count-comparison}), performance remains relatively stable across time intervals.
Qwen3-0.6B consistently achieves the lowest MAE across nearly all modality and interval combinations, with errors ranging from 0.51 to 0.60. Qwen3-1.7B performs similarly but with slightly higher error, while Granite-3.3-2b exhibits the weakest performance, particularly for subjective inputs where MAE reaches 0.86 at 2W.
Objective inputs generally provide the most reliable predictions, while subjective inputs are more variable across models. The combined modality offers modest improvements in some longer-horizon settings, especially for Granite.
Overall, Qwen3-0.6B is the strongest and most consistent model under NLS prompting, indicating that model selection has a greater impact on performance than time interval length.

\subsection{Model Accuracy: Statistical Summary}
Table \ref{tab:ssMAE} presents model performance under the SS prompting strategy.
Across all temporal windows, the Qwen3 models consistently outperform Granite-3.3-2B, particularly when objective features are used.
The lowest MAE overall is achieved by Qwen3-1.7B using objective features at the 1D interval (0.43 $\pm$ 0.24).
Subjective features generally produce comparable performance for the Qwen models, with Qwen3-0.6B achieving the best subjective results across all temporal windows (0.48--0.52 MAE).
Objective + subjective fusion performs similarly to the strongest single-modality results and yields the lowest errors at the 1M interval for both Qwen models (0.48--0.49 MAE), suggesting that statistical summaries enable effective multimodal integration over longer time intervals.
In contrast, Granite-3.3-2B exhibits consistently higher error, particularly for subjective inputs, indicating reduced ability to leverage self-reported features under the SS representation.
Overall, the results demonstrate that compact Qwen models are more effective than Granite-3.3-2B at utilizing statistical summaries for stress estimation.

\subsection{On-device Performance Evaluation}
We evaluate on-device performance using two core system metrics: end-to-end latency and generation throughput (Figure~\ref{fig:on-device-eval}).
Together, these results characterize the practical performance envelope of lightweight LLMs under realistic mobile workloads.

\subsubsection{Latency}
Latency exhibits clear differences across models and temporal intervals.
Qwen3-0.6B consistently achieves sub-second latency (0.37--0.47\,s) with minimal variation.
Qwen3-1.7B remains below 1.2\,s for 1D--2W intervals but increases to 2.43\,s at 1M..
Granite-3.3-2B exhibits the highest latency, peaking at 3.18\,s (1W) and 3.00\,s (2W).
Overall, Qwen3-0.6B shows the most stable latency profile, while larger models are more sensitive to temporal interval length.

\subsubsection{Throughput}
Throughput generally decreases with model size.
Qwen3-0.6B sustains the highest generation rates (78.0--82.9 tokens/s), followed by Qwen3-1.7B (51.1--56.9 tokens/s) and Granite-3.3-2B (32.0--36.3 tokens/s).
Across all models, throughput varies only modestly across temporal intervals, indicating consistent generation performance over time.

\section{Discussion}
Our results yield several insights with implications for mobile health system design, temporal modeling of mental states, and the practical deployment of LLMs under mobile resource constraints.

\subsection{Traditional ML vs. Zero-Shot ODLMs}
The performance gap between traditional supervised models and zero-shot ODLMs is relatively small.
While the best supervised baselines (SVM and RF) achieve the lowest MAE values of 0.50 across several configurations, the best-performing ODLM, Qwen3-0.6B-4bit, achieves a comparable mean MAE of 0.51 (SS).
This comparison should be interpreted in light of the different learning paradigms.
Traditional ML models are trained directly on labeled patient data and may require retraining when deployed to new populations or settings.
In contrast, ODLMs operate entirely in a zero-shot manner, requiring no patient-specific training and remaining readily deployable across novel users and contexts.
This flexibility is particularly valuable in mobile health applications, where labeled data are often limited, costly to obtain, and subject to privacy constraints.

\subsection{Natural Language String vs. Statistical Summary}
SS consistently outperforms
NLS, achieving lower MAE across all modalities and forecast horizons.
Unlike NLS, which requires models to process increasingly long serialized time-series sequences, SS provides a compact and fixed-format representation of feature distributions.
Consequently, NLS exhibits greater variability across horizons, whereas SS maintains stable and often improved performance, particularly at longer horizons.
These findings suggest that reducing prompt complexity is more beneficial than preserving raw sequential detail for long-range forecasting.

\subsection{Impact of Multimodal Fusion}
Multimodal fusion yields limited and inconsistent improvements across models and time intervals.
In the NLS setting (Table \ref{tab:nlsMae}), combined inputs generally perform similarly to the best single modality, with the most notable improvement observed for Granite-3.3-2b at the 1M horizon.
A similar pattern is seen for the SS representation (Table \ref{tab:ssMAE}), where fusion occasionally improves performance but rarely outperforms the strongest unimodal input.
Overall, the results suggest that multimodal fusion offers modest, model-dependent benefits, indicating that the current fusion strategy may not fully exploit complementary information from both modalities.

\subsection{Limitations}
Several limitations of this work should be acknowledged.
First, all evaluations are  conducted using zero-shot prompting without fine-tuning; domain-adapted models may  substantially reduce the accuracy gap observed relative to supervised baselines.
Second, experiments are conducted on a single dataset
and results may not generalize across populations, devices, or stress measurement scales.
Third, device-level evaluations span multiple iPhone generations
and per-device performance breakdowns were not reported, which may obscure hardware-specific variation.
Finally, energy consumption was not systematically measured, representing an important gap for battery-constrained deployments that warrants dedicated evaluation in future work.

\section{Conclusion}
This work demonstrates the feasibility of zero-shot stress prediction using compact, quantized ODLMs on mobile devices.
Despite requiring no patient-specific training, the smallest evaluated model, Qwen3-0.6B-4bit, achieves competitive accuracy alongside sub-second latency and a minimal memory footprint; underscoring generalizability and efficiency as key advantages over traditional supervised approaches in privacy-sensitive mobile health settings.

\bibliographystyle{ieeetr}
\bibliography{bibs/final}

@inproceedings{wang2024efficient,
  title={Efficient and personalized mobile health event prediction via small language models},
  author={Wang, Xin and Dang, Ting and Kostakos, Vassilis and Jia, Hong},
  booktitle={Proceedings of the 30th Annual International Conference on Mobile Computing and Networking},
  pages={2353--2358},
  year={2024}
}

@article{nissen2025medicine,
  title={Medicine on the Edge: Comparative Performance Analysis of On-Device LLMs for Clinical Reasoning},
  author={Nissen, Leon and Zagar, Philipp and Ravi, Vishnu and Zahedivash, Aydin and Reimer, Lara Marie and Jonas, Stephan and Aalami, Oliver and Schmiedmayer, Paul},
  journal={arXiv preprint arXiv:2502.08954},
  year={2025}
}

@article{kim2024healthllm,
  title={Health-llm: Large language models for health prediction via wearable sensor data},
  author={Kim, Yubin and Xu, Xuhai and McDuff, Daniel and Breazeal, Cynthia and Park, Hae Won},
  journal={arXiv preprint arXiv:2401.06866},
  year={2024}
}

@inproceedings{thambawita2020pmdata,
  title={Pmdata: a sports logging dataset},
  author={Thambawita, Vajira and Hicks, Steven Alexander and Borgli, Hanna and Stensland, H{\aa}kon Kvale and Jha, Debesh and Svensen, Martin Kristoffer and Pettersen, Svein-Arne and Johansen, Dag and Johansen, H{\aa}vard Dagenborg and Pettersen, Susann Dahl and others},
  booktitle={Proceedings of the 11th ACM Multimedia Systems Conference},
  pages={231--236},
  year={2020}
}

@article{gruver2023large,
  title={Large language models are zero-shot time series forecasters},
  author={Gruver, Nate and Finzi, Marc and Qiu, Shikai and Wilson, Andrew G},
  journal={Advances in Neural Information Processing Systems},
  volume={36},
  pages={19622--19635},
  year={2023}
}

@article{jin2023time,
  title={Time-llm: Time series forecasting by reprogramming large language models},
  author={Jin, Ming and Wang, Shiyu and Ma, Lintao and Chu, Zhixuan and Zhang, James Y and Shi, Xiaoming and Chen, Pin-Yu and Liang, Yuxuan and Li, Yuan-Fang and Pan, Shirui and others},
  journal={arXiv preprint arXiv:2310.01728},
  year={2023}
}

@article{xu2022globem,
  title={GLOBEM dataset: multi-year datasets for longitudinal human behavior modeling generalization},
  author={Xu, Xuhai and Zhang, Han and Sefidgar, Yasaman and Ren, Yiyi and Liu, Xin and Seo, Woosuk and Brown, Jennifer and Kuehn, Kevin and Merrill, Mike and Nurius, Paula and others},
  journal={Advances in neural information processing systems},
  volume={35},
  pages={24655--24692},
  year={2022}
}

@article{bondarenko2021wquantization,
  title={Understanding and overcoming the challenges of efficient transformer quantization},
  author={Bondarenko, Yelysei and Nagel, Markus and Blankevoort, Tijmen},
  journal={arXiv preprint arXiv:2109.12948},
  year={2021}
}

@article{li2020pruning,
  title={Efficient transformer-based large scale language representations using hardware-friendly block structured pruning},
  author={Li, Bingbing and Kong, Zhenglun and Zhang, Tianyun and Li, Ji and Li, Zhengang and Liu, Hang and Ding, Caiwen},
  journal={arXiv preprint arXiv:2009.08065},
  year={2020}
}

@article{ji2021distribution,
  title={On the distribution, sparsity, and inference-time quantization of attention values in transformers},
  author={Ji, Tianchu and Jain, Shraddhan and Ferdman, Michael and Milder, Peter and Schwartz, H Andrew and Balasubramanian, Niranjan},
  journal={arXiv preprint arXiv:2106.01335},
  year={2021}
}

@inproceedings{thapa2025stressllm,
  title={StressLLM: Large Language Models for Stress Prediction via Wearable Sensor Data},
  author={Thapa, Bishal and Rivas, Micaela and Griffith, Henry and Rathore, Heena},
  booktitle={2025 IEEE International Conference on Consumer Electronics (ICCE)},
  pages={1--6},
  year={2025},
  organization={IEEE}
}

@inproceedings{belyaeva2023multimodal,
  title={Multimodal llms for health grounded in individual-specific data},
  author={Belyaeva, Anastasiya and Cosentino, Justin and Hormozdiari, Farhad and Eswaran, Krish and Shetty, Shravya and Corrado, Greg and Carroll, Andrew and McLean, Cory Y and Furlotte, Nicholas A},
  booktitle={Workshop on Machine Learning for Multimodal Healthcare Data},
  pages={86--102},
  year={2023},
  organization={Springer}
}

@article{donawa2024designing,
  title={Designing Survey-Based Mobile Interfaces for Rural Patients With Cancer Using Apple’s ResearchKit and CareKit: Usability Study},
  author={Donawa, Alyssa and Powell, Christian and Wang, Rong and Chih, Ming-Yuan and Patel, Reema and Zinner, Ralph and Aronoff-Spencer, Eliah and Baker, Corey E and others},
  journal={JMIR Formative Research},
  volume={8},
  number={1},
  pages={e57801},
  year={2024},
  publisher={JMIR Publications Inc., Toronto, Canada}
}

@article{wei2025mophes,
  title={MoPHES: Leveraging on-device LLMs as Agent for Mobile Psychological Health Evaluation and Support},
  author={Wei, Xun and Zhou, Pukai and Wang, Zeyu},
  journal={arXiv preprint arXiv:2510.16085},
  year={2025}
}

@inproceedings{ma2024understanding,
  title={Understanding the benefits and challenges of using large language model-based conversational agents for mental well-being support},
  author={Ma, Zilin and Mei, Yiyang and Su, Zhaoyuan},
  booktitle={AMIA Annual Symposium Proceedings},
  volume={2023},
  pages={1105},
  year={2024}
}

@article{liu2023chatcounselor,
  title={Chatcounselor: A large language models for mental health support},
  author={Liu, June M and Li, Donghao and Cao, He and Ren, Tianhe and Liao, Zeyi and Wu, Jiamin},
  journal={arXiv preprint arXiv:2309.15461},
  year={2023}
}

@article{zheng2023building,
  title={Building emotional support chatbots in the era of llms},
  author={Zheng, Zhonghua and Liao, Lizi and Deng, Yang and Nie, Liqiang},
  journal={arXiv preprint arXiv:2308.11584},
  year={2023}
}

@article{haque2024state,
  title={State-of-the-art of stress prediction from heart rate variability using artificial intelligence},
  author={Haque, Yeaminul and Zawad, Rahat Shahriar and Rony, Chowdhury Saleh Ahmed and Al Banna, Hasan and Ghosh, Tapotosh and Kaiser, M Shamim and Mahmud, Mufti},
  journal={Cognitive Computation},
  volume={16},
  number={2},
  pages={455--481},
  year={2024},
  publisher={Springer}
}

@inproceedings{dongre2024physiology,
  title={Physiology-driven empathic large language models (EmLLMs) for mental health support},
  author={Dongre, Poorvesh},
  booktitle={Extended Abstracts of the CHI Conference on Human Factors in Computing Systems},
  pages={1--5},
  year={2024}
}

@article{mcewen_revisiting_2020,
	title = {Revisiting the {Stress} {Concept}: {Implications} for {Affective} {Disorders}},
	volume = {40},
	copyright = {https://creativecommons.org/licenses/by-nc-sa/4.0/},
	issn = {0270-6474, 1529-2401},
	shorttitle = {Revisiting the {Stress} {Concept}},
	url = {https://www.jneurosci.org/lookup/doi/10.1523/JNEUROSCI.0733-19.2019},
	doi = {10.1523/JNEUROSCI.0733-19.2019},
	language = {en},
	number = {1},
	urldate = {2026-02-20},
	journal = {The Journal of Neuroscience},
	author = {McEwen, Bruce S. and Akil, Huda},
	month = jan,
	year = {2020},
	pages = {12--21},
}

@article{hua2025charting,
  title={Charting the evolution of artificial intelligence mental health chatbots from rule-based systems to large language models: a systematic review},
  author={Hua, Yining and Siddals, Steve and Ma, Zilin and Galatzer-Levy, Isaac and Xia, Winna and Hau, Christine and Na, Hongbin and Flathers, Matthew and Linardon, Jake and Ayubcha, Cyrus and others},
  journal={World Psychiatry},
  volume={24},
  number={3},
  pages={383--394},
  year={2025},
  publisher={Wiley Online Library}
}

\end{document}